\documentclass[letterpaper, 10 pt, conference]{ieeeconf}  

\IEEEoverridecommandlockouts                              

\usepackage{amssymb, mathtools}
\usepackage{algorithm}
\usepackage[keeplastbox]{flushend}
\usepackage[noend]{algpseudocode}
\usepackage{xcolor}
\usepackage{booktabs}
\usepackage{siunitx}
\usepackage{cleveref}

\usepackage[backend=bibtex,natbib=true,style=ieee,mincitenames=1,maxcitenames=2,maxbibnames=20,url=false,doi=false]{biblatex}
\title{\LARGE \bf
Deep Binary Reinforcement Learning for Scalable Verification
}

\author{Christopher Lazarus$^{1}$ and Mykel J. Kochenderfer$^{2}$
}

\begin{document}

\maketitle
\thispagestyle{empty}
\pagestyle{empty}

\begin{abstract}

The use of neural networks as function approximators has enabled many advances in reinforcement learning (RL). The generalization power of neural networks combined with advances in RL algorithms has reignited the field of artificial intelligence. Despite their power, neural networks are considered black boxes, and their use in safety-critical settings remains a challenge. Recently, neural network verification has emerged as a way to certify safety properties of networks. Verification is a hard problem, and it is difficult to scale to large networks such as the ones used in deep reinforcement learning.
We provide an approach to train RL policies that are more easily verifiable. We use binarized neural networks (BNNs), a type of network with mostly binary parameters. We present an RL algorithm tailored specifically for BNNs. After training BNNs for the Atari environments, we verify robustness properties.

\end{abstract}

\section{INTRODUCTION}

In reinforcement learning (RL), an agent observes the state of the environment and acts accordingly to maximize the expected accumulation of reward. RL has been applied to many different domains ranging from clinical \cite{rl-sepsis} to financial decision making \cite{kolm2020modern, lim2020time}.
There is growing interest in using RL to optimize safety critical systems, such as autonomous vehicles \cite{Julian2019, xiang2018verification}.


Contemporary RL approaches \cite{dqn-human} often rely on deep neural networks (DNNs) to represent the policy that governs the agent's behavior. DNNs are flexible and powerful approximators but are also considered black-boxes as it is hard for humans to characterize, let alone parse their transfer function. 
Our concern is diagnosing whether a policy represented by a neural network will induce safe behavior. We approach this issue by certifying input-output properties of the neural network itself, taking advantage of recent advances in neural network verification \cite{bnn-survey}.

A common issue with current neural network verification approaches is that, given the hardness of the verification problem, algorithms do not scale to large networks such as those widely used in RL.  \Citet{katz2017reluplex} show that verification of ReLU networks is NP-hard. In general, verification of a network with $n$ nodes can require an exponential amount of work, as $2^n$ potential activation patterns could have to be considered. Some approaches, including the one we use, rely on mixed-integer programming (MIP) which is also known to be NP-hard.

Recent work has shown that BNN verification scales better \cite{lazarus2021mixed, Narodytska2020In}, and BNNs are able to attain similar performance to full precision neural networks in computer vision classification tasks \cite{bnn-ste, bnn-survey} . We present a method to solve deep RL problems with binarized networks which produces more easily verifiable policies.

Our contributions are: 
\begin{itemize}
	\item A framework for training action-value functions using binarized neural networks that attain similar performance to full precision counterparts with a significantly smaller memory footprint and computational cost.
	\item The binarized nature of the networks enables more scalable verification for safety-critical scenarios.
	\item A demonstration of the capability to evaluate the robustness of the policies using existing BNN verification approaches as well as the certification of safety properties that scale far beyond those of full precision networks.
\end{itemize}
These contributions combined pave the way for the use of deep reinforcement learning in safety-critical settings, attaining similar performance to methods that rely on neural networks for function approximation while simultaneously allowing verifiability.

\section{BACKGROUND}

Our contributions are built upon previous advances in reinforcement learning, model compression, and neural network verification.
This section provides a brief overview of these topics.

\subsection{Reinforcement Learning}
Markov Decision Processes (MDPs) have been used to model autonomous vehicles such as self-driving cars \cite{Bouton2020} and unmanned aircraft systems \cite{Kochenderfer2015, Temizer2010}. MDPs are a powerful framework for modeling sequential decision-making under uncertainty problems where an agent has to navigate the world and seeks to maximize the reward it collects.

An MDP is composed of a state space $S$, an action space $A$, which is the set of possible actions that the agent can take at each step, a transition function $T$ that specifies the probability of arriving at the next state $s'$ given that action $a$ was taken by the agent at state $s$ and, a reward function $R$ that specifies the reward received when transitioning from state $s$ by performing action $a$. 

A policy $\pi(s)$ is a function that determines which action $a$ to perform at each state $s$. 
The utility of following policy $\pi$ from state $s$ is denoted $U^\pi(s)$ and is typically referred to as the value function. 
Solving an MDP corresponds to crafting a policy $\pi^*$ that leads to maximizing the expected utility $\mathbb{E} \left [ U^\pi \right ]$:
\begin{align}
	\pi^*(s) = \arg \max_\pi U^\pi (s)
\end{align}

When all the elements that define an MDP are known explicitly, then it is possible to solve the MDP exactly using dynamic programming methods \cite{Kochenderfer2015} . 
However, in many interesting cases, the transition function is not accessible, but reinforcement learning can be used to optimize a policy.

\subsection{Deep $Q$-Learning}
A framework for training agents directly from high-dimensional sensory inputs like images using RL has been demonstrated to attain super-human performance \cite{dqn}. 
RL faces several challenges that are compounded when using deep learning. Many deep-learning applications require large amounts of labeled data. 
On the other hand, RL agents have to learn from a scalar reward signal that is often noisy, sparse, and delayed. 
Another issue is that many deep learning algorithms assume that the data is independent and identically distributed; in RL, it is typical to encounter sequences of highly correlated observations. 
Furthermore, the distribution of the observations changes as the agent learns to navigate the environment, which further undermines the assumption of a fixed underlying distribution.

A solution to many of these challenges relies on a variant of the $Q$-learning algorithm \cite{Watkins92q-learning}. 
$Q$-learning is a model-free reinforcement learning algorithm used to estimate the state-action value function $Q$ and involves applying incremental updates to estimates of the Bellman Equation:
\begin{align}
	Q(s,a) &= R(s,a) + \gamma \sum_{s'} T( s' \mid s,a) U(s')\\
	&= R(s,a) + \gamma \sum_{s'} T( s' \mid s,a) \max_a Q(s',a')
\end{align}
Instead of using the transition function $T$ and the reward function $R$, we use the observed state $s$, next state $s'$ and reward $r$ obtained after performing action $a$.

The key idea behind many reinforcement learning algorithms is to iteratively estimate the state-action value function by using the Bellman equation. 
In practice, this approach is often not viable because the value function is estimated separately for each experience tuple without the possibility for generalization. 
An alternative is to use function approximation to estimate the function $Q(s,a; \theta) \approx Q^*(s,a)$. This approximation can made using a linear function, which has the advantage that it is easier to verify. 

In Deep $Q$-learning \cite{dqn}, a network is trained with stochastic gradient descent to update the weights. 
To address the problem of correlated data and non-stationary distributions, an experience replay mechanism is used, which randomly samples over previous experience tuples. 
\Citet{dqn-human} modified the $Q$-learning algorithm to help stabilize learning. 
A duplicate of the network is used and updated at a slower pace than the policy network.

In the original DQN paper \cite{dqn}, a preprocessing function $\phi$ was introduced. The function $\phi$ downsamples and crops the input space from the original $210 \times 160$ pixels with a 128 color palette to $84 \times 84 \times 4$ grayscale. 
The last dimension is introduced to concatenate $4$ consecutive frames to induce the Markovian property over the environment, given that motion cannot be captured by a single frame. 
Furthermore, the DQN framework uses the frame-skipping technique by repeating the same action for a sequence of steps.

Despite parameter sharing, the network's architecture requires a large number of parameters, which has, so far, precluded the verification of similar-sized convolutional networks. 
This problem is the motivation behind our use of binarized neural networks.

\subsection{Neural Network Verification}
Neural networks are susceptible to adversarial attacks \cite{papernot2016limitations}, producing drastically different outputs when slightly perturbing their input which in the context of RL can lead to unsafe policies.
Multiple approaches to evaluate the robustness of networks to adversarial attacks have been developed. 
These only provide statistical assessments and focus on evaluating the network on a large but finite collection of points in the input space. 
However, given that the input space is, in principle, infinite in cardinality, it is not viable to assess the output for all the points in the input space.
Recently, new approaches have emerged as an alternative to formally certify the input-output properties of neural networks \cite{OPT-035}. 

The verification problem consists of checking whether input-output relationships of a function hold \cite{OPT-035}. A subset of the input space $\mathcal{X} \subseteq D_x$ and a subset of the output space $\mathcal Y \subseteq D_y$ are defined. 
Solving the verification problem requires certifying whether the following holds:
\begin{align}
	x \in \mathcal{X} \implies y = F(x) \in \mathcal{Y}
\end{align}
In general, the input and output sets could have any geometry but are often constrained by the verification algorithm. 

Verification is a complex problem, and current algorithms can only terminate with small and relatively simple networks. 
Networks with convolutional layers pose a challenge to verification algorithms, making verifying policy networks such as those used in DQN a very difficult task. 

\subsection{Binarized Neural Networks}
With the goal of training a policy that uses neural networks but is more easily verified, we decided to use simpler models. 
Binarized Neural Networks perform well at vision tasks \cite{bnn-survey} and provide significant reductions in memory footprint, computational cost, and inference speed while attaining similar performance \cite{bnn-steDEP, surveyBNN1}.

A binarized neural network is a network involving binary weights, and activations \cite{bnn-ste}. 
The goal of network binarization is to represent the floating-point weights $W$ and the activations $z_{i,j}$ for a given layer using $1$ bit. 
The parameters are represented by:
\begin{align}
	Q(W) = \alpha B_{W} && Q(z) = \beta B_{z}
\end{align}
where $B_{W}$ and $B_{z}$ are binarized weights and binarized activations, with optional scaling factors $\alpha$ and $\beta$ used for batch normalization.
The $\operatorname{sign}$ function is often used to compute $Q_{W}$ and $Q_{z}$:
\begin{equation}
	\operatorname{sign} (x) = \left\{\begin{array}{ll}
+1,  &\text{if } x \geq 0 \\ 
-1,  &\text{otherwise }
\end{array}\right.
\end{equation}
This structure enables easy implementation of batch normalization while keeping most parameters and operations binary.

In this context, the arithmetic operations needed for a forward pass of a layer $z^b$ in a binarized network $F^b$ can be computed as:
\begin{align}
	z^b_i &= \sigma \left (Q(W) z^b_i \right ) = f_i (z^b_{i-1}) = \sigma_i \left ( Q(W)_i z^b_{i-1} \right )\\
	&= \sigma \left ( \alpha \beta B_W \circledcirc B_z \right ) = \alpha \beta \operatorname{sign}\left ( B_W \circledcirc B_z \right ) 
\end{align}
where $\circledcirc$ denotes the inner product for binary vectors with bitwise operation XNOR-Bitcount, leading to efficient ad-hoc hardware implementations of the networks for inference.

Binarization introduces non-differentiable and even non-continuous blocks to the computational graph, complicating the optimization used to train the network. 
Recent work motivated by their applicability in highly constrained environments such as edge devices has enabled them to achieve performance comparable to traditional full precision networks \cite{bnn-ste}.

The reduced memory requirement and simplified computation resulting from this representation have a drawback: binarized neural networks are harder to train. 
A significant challenge is back-propagating the gradient of the weights through sign functions. 
There are workarounds \cite{bnn-survey} such as using straight-through estimators (STE) \cite{bnn-ste}. 
This challenge originally identified for image classification could also significantly hinder the performance of BNNs for RL tasks, particularly if the controller's inputs are pixels, as in the case of the original DQN work.

\subsection{Verification of Binarized Neural Networks}
Because we use BNNs for value function approximation, our verification algorithm can exploit the simplified structure of the networks. 
BNNs are composed of piecewise-linear layers that may have piecewise-linear activation functions, such as ReLU, $\max$, and $\operatorname{sign}$.

There are multiple approaches to efficiently verify BNNs. 
Some approaches rely on SAT solvers and reduce the verification problem to a Boolean satisfiability problem \cite{narodytska2017verifying, Narodytska2020In}, which limits their applicability exclusively to fully binarized networks with binary inputs and outputs. 
Another SAT solver-based approach that can handle BNN constraints to speed up the verification process was introduced \cite{kai}. 
One SMT-based approach that can be applied to both binary and full precision networks is a technique that extends the Reluplex \cite{katz2017reluplex} algorithm by including various BNN specific optimizations an was introduced by \citet{amir2021smtbased}.

Given that our networks take non-binary inputs, we will verify our networks with the simple mixed-integer programming (MIP) introduced by \citet{lazarus2021mixed}. 
Each component of the network is encoded as a set of linear equations that describe the forward-pass of the network. 
The input set $\mathcal{X}$ is also encoded as a set of linear constraints, which implicitly limits the sets that can be represented to intersections of halfspaces. 
Finally, $\mathcal{Y}^C$, the complement of the output set $\mathcal{Y}$ is encoded with the corresponding variables associated with the output layer of the network. 
A MIP solver is used to search for a feasible assignment. If a feasible assignment is found, it corresponds to a counterexample of the desired property. 
If no feasible assignment is found, then we can conclude that $F(\mathcal{X}) \subseteq  \mathcal{Y}$ and the property holds.
MIP verification approaches can verify BNNs with more nodes than full-precision networks and offer thus an option to verify BNN based RL policies.

\section{Deep Binary $Q$-Learning}
We aim to train BNNs that approximate the optimal $Q$-function for the given environments. 
We base our approach on DQN \cite{dqn-human}. However, our initial results showed that replacing the networks described by \citet{dqn} with binarized versions results in significant performance degradation. 
A specific approach tailored for BNNs is needed.

In DQN, the network takes a preprocessed image from the Atari game environment $\phi(x)$ as the input and outputs a vector of $Q$-values for each action. 
The preprocessed inputs summarize the state of the environment $s$, and each output $\hat{q}(a)$ corresponds to the estimated value for action $a$. 

Our initial approach was to replace the original network $\hat{q}$ with a binarized network $\hat{q}_b$ with the same architecture but with all parameters binarized and a scaling factor for each filter and fully connected layer. 
This approach also replaced the target network. Our initial evaluation found that the networks failed to converge, and the performance degraded significantly. 

To address the issues that arise with BNNs, we propose two modifications. 
The first is to use a larger network. A BNN with the same architecture as a DNN has less expressivity but can still be computed efficiently and stored with approximately $32 \times$ less memory. 
Adding layers to a BNN can increase expressivity but still produce a network with a smaller memory footprint than the DNN counterpart. 
The second modification is to borrow ideas from transfer learning in model compression and keep the target network as a DNN using the full precision parameters that are used as part of the training process for the binarized network. 
We use the output of this higher resolution target network to compute the loss and gradient updates for the binarized value network.
Memory replay was left unmodified. \Cref{alg:q-learning} outlines the approach.

\begin{algorithm}
  \caption{Binary $Q$-Learning}\label{alg:q-learning}
  \begin{algorithmic}[1]
    \State{Populate replay buffer $D$ to capacity}
      \State Initialize targrt action-value function $\hat{q}_t$ with random $w^-$
      \State Initialize action-value function $\hat{q}$ with $w=\text{sign}(w^-)$
      \For{episode $m=1,\dots,M$}
		\State Select action $a_t = \begin{cases}
 & \text{ random }  \text{with probability } \epsilon \\
 & \arg \max_a \hat{q}(s_t, a, w)) \text{ o.w. }
\end{cases}$ 
        \State Execute $a_t$ and observe $r_t$ and $x_{t+1}$
        \State Compute $s_{t+1}$ with $s_t$ and $x_{t+1}$
        \State Insert transition $(s_t, a_t, r_t, s_{t+1})$ in $D$
        \State Sample random minibatch of $N$ transitions from $D$
        \State Set $y_j = r_j$ if episode ends at step $j+1$
        \State otherwise set $y_j = r_j + \gamma \max_{a'} \hat{q_t}(s_{j+1}, a', w^-)$
        \State SGD step on $J(w) = \frac{1}{N}\sum_{j=1}^{N} (y_j -\hat{q}(s_j, a_j, w))^2$
        \State Every $C$ steps set $w^- = \text{sign}(w)$
      \EndFor
  \end{algorithmic}
\end{algorithm}

\section{Experiments}
The ultimate goal is to produce verifiable policies optimized using RL. We begin by training the policies using Deep Binary $Q$-Learning, and later we verify some of their properties. 
Our experiments are accordingly divided into two sets.
The first set of experiments demonstrates the use of BNNs for RL tasks in the Atari environment. 
The second set of experiments demonstrates the capability of verifying the policies using MIP.

The original DQN network has an input of an $84 \times 84 \times 4$ image; the first convolutional layer has $32$ filters of size $8 \times 8$ with stride $4$ and is followed by a ReLU nonlinearity. The second hidden layer is composed of $64$ filters of size $4 \times 4$ with stride $2$, again followed by a ReLU. 
The third convolutional layer has $64$ filters of $3 \times 3$ with stride $1$ and another ReLU. The last hidden layer is fully connected with $512$ ReLUs, and the output layer is fully connected.

For all our experiments, we used a network with the original DQN architecture, a BQN network with the same architecture, and another larger binarized network with two more convolutional layers labeled BQN-L.


\subsection{Deep Binary $Q$-Learning for Atari}
We trained agents on the seven Atari games evaluated with DQN \cite{dqn}: Beam Rider, Breakout, Enduro, Pong, Q*bert, Seaquest, and Space Invaders. We trained full precision DQNs with the same network architecture, learning algorithm, and hyperparameters for all games. The reward function was modified for training so that all positive rewards are $1$ and all negative rewards $-1$, as in the original DQN approach. 
Aiming to benchmark our approach against DQN, we used stochastic gradient descent (SGD) with root mean squared propagation (RMSProp) with minibatches of size $32$. 
During training, an $\epsilon$-greedy policy was used with a linear decay from $1.0$ to $0.1$ for the first million steps.

We measured the performance of the agents using the same approach as \citet{dqn-human} by recording the mean score attained when running an $\epsilon$-greedy policy with $\epsilon = 0.05$. 
We include the scores reported for human players (median score after two hours of play) reported originally by \citet{dqn}.

\begin{table}[]
\caption{Comparison of average total reward for the different learning methods}
\begin{center}
\label{table:rl-res}
\begin{tabular}{@{}lrrrrrr@{}}
\toprule
            & RNG & Human & DQN  & B-DQN & BQN & BQN-L \\ \midrule
B. Rider    & 356    & 7456  & 4113 & 401   & 807 & 4019  \\
Breakout    & 1.1    & 31    & 163  & 2.8   & 67  & 156   \\
Enduro      & 0      & 368   & 501  & 0     & 33  & 488   \\
Pong        & $-$21.1  & $-$3    & 19   & $-$9    & $-$4  & 18    \\
Q*Bert      & 148    & 18900 & 2017 & 179   & 498 & 2033  \\
Seaquest    & 117    & 28010 & 1707 & 133   & 872 & 1698  \\
S. Invaders & 163    & 3690  & 603  & 192   & 353 & 589   \\ \bottomrule
\end{tabular}
\end{center}
\end{table}

Our initial attempt at training a BQN by substituting the value and target networks in the DQN framework with binarized networks failed. 
We performed a simple ablation study and identified that using our modified algorithm significantly improved the average collected rewards.

Our results are captured in \cref{table:rl-res}. RNG is a random policy. 
The human average rewards are the ones reported by \citet{dqn}. 
DQN is the policy we trained following the original framework. 
B-DQN is a binarized network with the same architecture as the original network using the unmodified DQN algorithm. 
BQN is our approach with a network of the same size as DQN. BQN-L is our approach with a network with two more convolutional layers.

It is clear that substituting the original network with a binarized version hindes performance significantly. 
Our algorithm observed improvement with the BQN network with the same architecture as the DQN network. 
After increasing the network size, we observed almost equal performance with the binarized network.
Recall that BQN is $\sim31 \times$ smaller than DQN and that BQN-L is $\sim21 \times$ smaller.

\subsection{Verification of BNN controllers}
Our second set of experiments consists of attempting to verify properties of the networks. 
In some contexts, the properties can have a clear meaning. For example, for a vision network, one might want to verify that an image with the color red is never classified as a go signal. 
In other settings, one could verify that the controller of a cyber-physical system does not lead the system to unsafe situations such as the properties verified by \citet{katz2019marabou}.

Analyzing the robustness of vision networks has attracted a lot of attention \cite{papernot2016limitations}. 
In the context of image classification, a network $F$ assigns a value to each of the possible labels in its training set, and the maximum value $\arg \max_i y_i$ is often used to impute the label of an input $x$. 
Despite having pixel inputs, the policies for RL are not trained to classify images but rather regress the $Q$-function of the MDP. 
However, the action is selected at each step by identifying the corresponding maximal value across all actions. We can analyze the robustness of the policies to conclude whether the optimal action performed by the agent would change give a small perturbation of the input frames.

We consider an input $x_0$ with associated action $a^{i^*} \in A$. It would be desirable that $y_{i^*} > y_j$ for all $j \neq i^*$, which can be encoded with the following sets:
\begin{align}
	\mathcal{X} &= \left \{ x \in D_x : \left \| x-x_0  \right \|_p \leq \epsilon \right \},\label{eq:robustness1}\\
	\mathcal{Y} &= \left \{ y \in D_y : y_{i^*} > y_j \forall j \neq i^* \right \},
\label{eq:robustness2}
\end{align}
where $\epsilon$ is the radius of the allowable perturbation in the input. 
If $p=1$ or $p= \infty$, we have linear constraints that can be easily encoded as part of the MIP formulation.
Encoding the output set $\mathcal{Y}$ is not possible with a single linear program, given that the maximum operator requires a disjunction of half-spaces. 
With our MIP formulation, the set can be encoded directly with the addition of dummy binary variables.

To assess the verifiability of the binarized networks, we designed simple properties to verify. 
We focused on the DQN, BQN, and BQN-L networks trained for the Q*bert environment, where the BQN-L network had the most competitive performance. 
We selected $50$ input frames and identified the optimal action $a^* = \arg \max_a Q(s,a)$ for each of those frames. We constructed an input set using the $\ell_1$ norm with a radius of $\epsilon = .01$ in the normalized grayscale, which corresponds to small perturbations in the luminance challenge of the images. 
We attempted to verify all the properties for the $3$ networks. For the DQN network, we used Marabou \cite{katz2019marabou}, and for the BQN networks, we used the MIP approach described by \citet{lazarus2021mixed}. We set a limit of $30$ minutes for each property to be verified.

\begin{table}[h]
\caption{Properties verified for each network.}
\label{table:verif}
\begin{center}
\begin{tabular}{@{}lrrr@{}}
\toprule
           & DQN & BQN & BQN-L \\ \midrule
Verified   & 0   & 39  & 26    \\ 
Unverified & 50 & 11  & 24    \\ \bottomrule
\end{tabular}
\end{center}
\end{table}

\Cref{table:verif} summarizes the results of our verification experiments. We were unable to verify any network properties with the original DQN architecture with full precision parameters within the timeout limit. For the BQN network, we were able to verify about $78\%$ of the properties. For the larger and better performing BQN-L, we were able to verify $52\%$ of the properties.

\section{CONCLUSIONS}

We identified an opportunity for end-to-end verifiable deep reinforcement learning in this work. 
The key idea was to use simpler models for value function approximation. 
We presented a framework to train binarized neural networks for reinforcement learning tasks. 
Our approach is inspired by the DQN algorithm but with specific modifications for binarized networks. 
The networks displayed competitive performance in the Atari environment for which they were trained. 
Furthermore, we demonstrated that binarized networks can be verified. 
Combining these two contributions can pave the way for the use of deep reinforcement learning in safety-critical settings.

Potential avenues for future work include investigating how binarized neural networks can be used in more sophisticated versions of value function-based RL algorithms such as the Double Deep $Q$-Network (DDQN) or the Dueling Network. 
Additionally, binarized neural networks could also be used in policy gradient methods. 
However, we suspect the non-differentiability of the binary blocks could make BNNs unsuitable for that kind of approach.

\addtolength{\textheight}{-12cm}   





\section*{ACKNOWLEDGMENT}
The NASA University Leadership initiative (grant \#80NSSC20M0163) provided funds to assist the authors with their research, but this article solely reflects the opinions and conclusions of its authors and not any NASA entity.



\renewcommand*{\bibfont}{\small}
\printbibliography

\end{document}